\documentclass[twoside]{article}

\usepackage[preprint]{aistats2022}

\usepackage[utf8]{inputenc}           
\usepackage[T1]{fontenc}              
\usepackage{hyperref}                 
\usepackage{url}                      
\usepackage{booktabs}                 
\usepackage{amsfonts,amsmath,amssymb} 
\usepackage{nicefrac}                 
\usepackage{microtype}                
\usepackage{xcolor}                   
\usepackage{tikz}
\usepackage{subcaption}

\usepackage{floatrow}
\usepackage{enumitem}
\usepackage{tabularx}
\usepackage{float}
\usepackage{hhline}
\usepackage{multirow}
\newcolumntype{Y}{>{\centering\arraybackslash}X}
\usepackage{comment}
\usepackage{multirow}

\newcommand{\bcket}[3]{\left#1 #3 \right#2}

\renewcommand{\b}{\bcket{(}{)}}

\newcommand{\cb}{\bcket{\{}{\}}}

\renewcommand{\P}[1][]{\operatorname{P}_{#1}\b}

\newcommand{\dd}[2][]{\frac{\partial #1}{\partial #2}}

\renewcommand{\ss}[1]{\{ #1 \}_{s=1}^S}

\newcommand{\tsum}{\textstyle{\sum}}

\newcommand{\Din}{\mathcal{D}_\text{in}}
\newcommand{\Dout}{\mathcal{D}_\text{out}}

\newcommand{\ceq}{\mkern2mu{=}\mkern1mu}
\newcommand{\bftab}{\fontseries{b}\selectfont}

\usepackage[round]{natbib}

\begin{document}

%

%

\twocolumn[

\aistatstitle{Bayesian OOD detection with aleatoric uncertainty and outlier exposure}

\aistatsauthor{ Xi Wang \And Laurence Aitchison }

\aistatsaddress{ University of Massachusetts Amherst \And  University of Bristol} ]

\begin{abstract}
Typical Bayesian approaches to OOD detection use epistemic uncertainty. Surprisingly from the Bayesian perspective, there are a number of methods that successfully use \textit{aleatoric} uncertainty to detect OOD points (e.g.\ Hendryks et al. 2018). In addition, it is difficult to use outlier exposure to improve a Bayesian OOD detection model, as it is not clear whether it is possible or desirable to increase posterior (epistemic) uncertainty at outlier points. We show that a generative model of data curation provides a principled account of aleatoric uncertainty for OOD detection. In particular, aleatoric uncertainty signals a specific type of OOD point: one without a well-defined class-label, and our model of data curation gives a likelihood for these points, giving us a mechanism for conditioning on outlier points and thus performing principled Bayesian outlier exposure. Our principled Bayesian approach, combining aleatoric and epistemic uncertainty with outlier exposure performs better than methods using aleatoric or epistemic alone.


\end{abstract}

\section{Introduction}\label{sec:intro}



The most typical approach to Bayesian OOD distribution detection uses epistemic uncertainty \citep{lakshminarayanan2017simple,malinin2018predictive, choi2018waic, wen2019batchensemble, malinin2020ensemble, postels2020hidden}.
We have epistemic uncertainty when finite training data fails to pin down the classifier's ideal outputs in all regions of the input space \citep{der2009aleatory,fox2011distinguishing,kendall2017what}.
Importantly, the amount of epistemic uncertainty will vary depending on how close a given test point is to the training data.
Close to the training data, the classifier's predictive distribution is reasonably well-pinned-down and there is little epistemic uncertainty.
In contrast, far from the training data, the classifier's predictive distribution is more uncertain, and this uncertainty can be used to detect OOD data.
In contrast, aleatoric uncertainty is the irreducible output ``noise'' that is left over when there is no uncertainty in the parameters (e.g.\ because a lot of training data is available).

Some work using Bayesian epistemic uncertainty for OOD detection explicitly rejects the use of aleatoric uncertainty \citep{malinin2018predictive,malinin2020ensemble,wen2019batchensemble,choi2018waic,postels2020hidden}, while other work implicitly combines aleatoric and epistemic uncertainty by looking at the overall predictive entropy \citep{lakshminarayanan2017simple,izmailov2021bayesian, ovadia2019can, maddox2019simple}.
Surprisingly from the Bayesian perspective, there are a large number of methods that successfully use aleatoric uncertainty alone to detect OOD points \citep{hendrycks2016baseline,liang2017principled,lee2017training,liu2018open,hendrycks2018deep}.
In addition, many of these methods can be \textit{trained} on OOD points, in a process known as outlier exposure (OE).

When suitable OOD data is available, OE leads to dramatic increases in performance \citep{hendrycks2018deep}.
However, developing a principled Bayesian method using OE is difficult.
In particular, a principled Bayesian formulation would involve treating the OE points as providing extra terms in the likelihood.
However, it not currently clear how to create a likelihood for outlier points.
Instead, current OE methods use a variety intuitively reasonable objectives, which have no interpretation as log-likelihoods and thus cannot be combined with Bayesian inference.

\begin{figure}[t]
    \centering
      \includegraphics[width=0.95\linewidth]{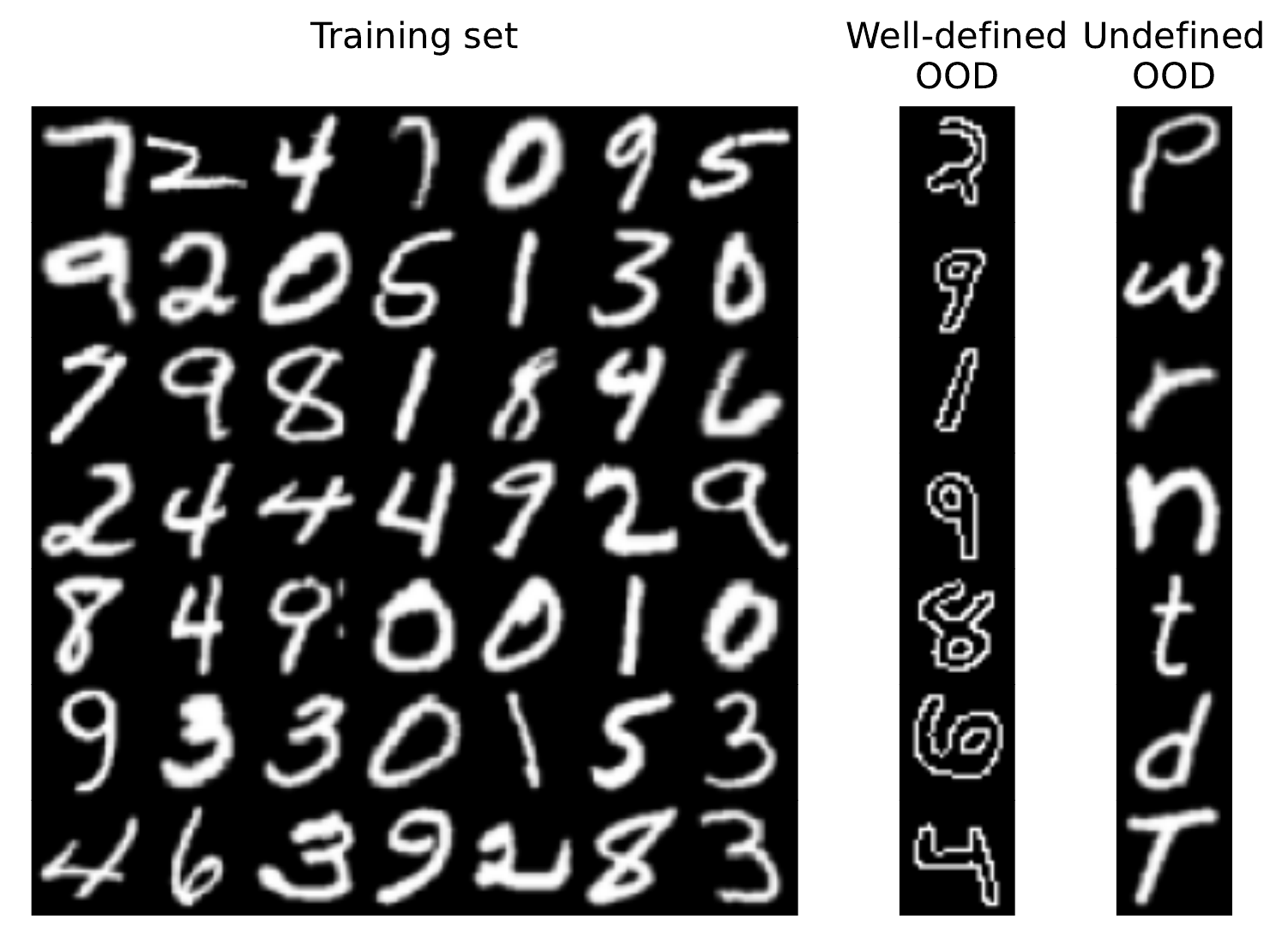}
    \caption{
      When training on MNIST (left), there is the potential for OOD images with a well-defined class-label (middle), and for images that simultaneously are OOD and have an undefined class-label (right).
      \label{fig:mnist}
    }
\end{figure}

In this paper, we provide a principled account of how to incorporate aleatoric uncertainty and outlier exposure into Bayesian OOD detection methods.
In particular, we consider a model of the curation process applied during the original creation of datasets such as CIFAR-10 and ImageNet.
Critically, this curation process is designed to filter out a specific set of OOD points: those without a well-defined class label. For simplicity, we will refer to these points as OOD for the remainder of the paper.
For instance, if we try to classify an image of a radio as cat vs dog, there is no well-defined class-label, and we should not include that image in the training set.
We model curation as a consensus-formation process. 
In particular, we give each the image to multiple human annotators: if the image has a well-defined label (Fig.~\ref{fig:mnist} left and middle), they will all agree, consensus will be reached and the datapoint will be included in the dataset.
In contrast, if the human annotators are given an image with an undefined class label (Fig.~\ref{fig:mnist} right), all they can do is to choose randomly, in which case they disagree, consensus will not be reached and the datapoint will be excluded from the dataset.
Critically, that random final choice corresponds to \textit{aleatoric}, not epistemic uncertainty.
If we ask a human to classify an image of a radio as cat vs dog, the issue certainly is not that the human annotator is uncertain about the radio's degree of ``cat-ness'' or ``dog-ness''.
The issue is that we are forcing the human to answer a fundamentally nonsensical question, and the only reasonable response is to choose randomly.
That random choice thus corresponds to \textit{aleatoric} uncertainty, and thus aleatoric uncertainty can signal that the point is OOD, and has an undefined class-label.
Our model gives a likelihood for being OOD (or having an ``undefined class label'') in terms of the underlying classifier probabilities, allowing us to incorporate outliers in principled Bayesian inference.
We find that our approach, incorporating OE and aleatoric uncertainty with Bayes performs better than a standard Bayesian approach without OE, and better than a standard aleatoric uncertainty based approach with OE \citep[e.g.][]{hendrycks2018deep}.

\begin{figure*}[t]
  \centering
  \begin{tikzpicture}
    \node[inner sep=0pt] (cat) at (0,0)
      {\includegraphics[height=2cm]{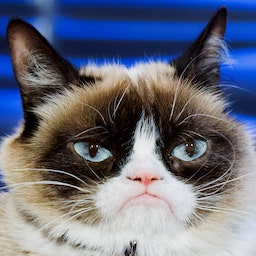}};
    \node[anchor=north] (lab) at (cat.south) {$Y_1=Y_2=Y_3=\text{cat}$};
    \node[anchor=north] (lab) at (lab.south) {$Y=\text{cat}$};
    
    \node[inner sep=0pt] (dog) at (4.5cm,0)
      {\includegraphics[height=2cm]{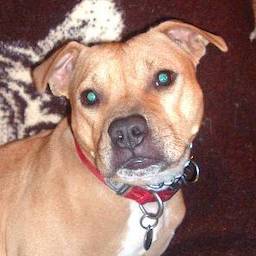}};
    \node[anchor=north] (lab) at (dog.south) {$Y_1=Y_2=Y_3=\text{dog}$};
    \node[anchor=north] (lab) at (lab.south) {$Y=\text{dog}$};
    \node[inner sep=0pt] (amb) at (9cm,0)
      {\includegraphics[height=2cm]{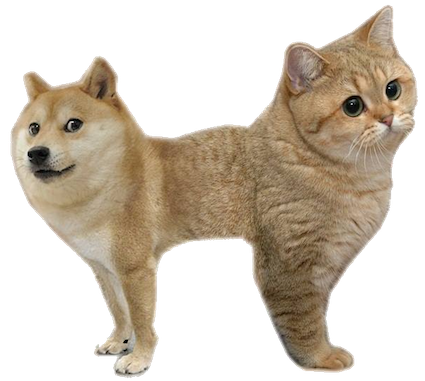}};
    \node[anchor=north] (lab) at (amb.south) {$Y_1=Y_2=\text{dog};\quad Y_3=\text{cat}$};
    \node[anchor=north] (lab) at (lab.south) {$Y=\texttt{Undef}$};
  \end{tikzpicture}
  \caption{
    An example of the generative model of dataset curation, with $S=3$.
    Three annotators are instructed to classify images as cats or dogs.
    The left and middle images have well-defined class-labels as they are clearly a cat and a dog respectively. 
    Therefore, the annotators all give the same answer and consensus is reached.
    However, the right-most image is OOD (it has an undefined class label), so the annotators are forced to give random answers; they disagree and we remove the image from the dataset.
    \label{fig:schematic}
  }
\end{figure*}
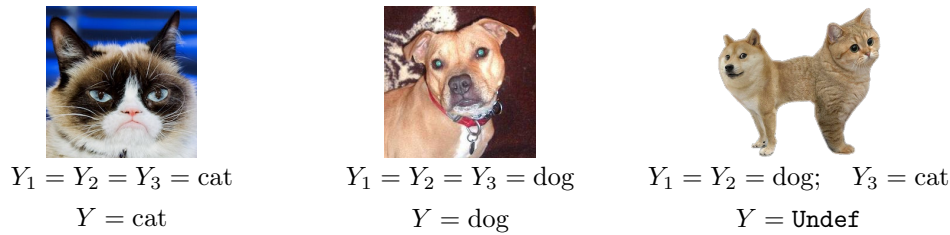

\section{Background}\label{sec:background}

In the introduction, we briefly noted that different annotators will agree about the class label when that class label is well-defined, but will disagree for OOD inputs without a well-defined class-label (if only because they are forced to the label the image and the only thing they can do is to choose randomly).
Interestingly, a simplified generative model which considers the probability of disagreement amongst multiple annotators has already been developed to describe the process of data curation \citep{aitchison2021statistical,aitchison2020statistical}.
In data curation, the goal is to exclude any OOD images to obtain a high-quality dataset containing images with well-defined and unambiguous class-labels.
Standard benchmark datasets in image classification have indeed been carefully curated.
For instance, in CIFAR-10, graduate student annotators were instructed that ``It's worse to include one that shouldn't be included than to exclude one'', then \citet{krizhevsky2009learning} ``personally verified every label submitted by the annotators''.
Similarly, when ImageNet was created, \citet{imagenet_cvpr09} made sure that a number of Amazon Mechanical Turk (AMT) annotators agreed upon the class before including an image in the dataset.

\begin{figure}
  \centering
  \begin{tikzpicture}
    \def\dx{1.5cm}
    \def\dy{0.7cm}
    \def\df{4cm}
    \def\da{-2cm}
    \node (A) at ({-0.5*\dx}, 0) {\textbf{A}};
    \node (X) at ({0}, 0) {$X$};
    \node (W) at ({0}, {-2*\dy}) {$\theta$};
    \node (Y) at ({\dx}, {-\dy}) {$Y$};
    \draw[->] (X) -- (Y);
    \draw[->] (W) -- (Y);
    
    \node (B) at ({\df-0.5*\dx}, 0) {\textbf{B}};
    \node (X) at ({\df}, 0) {$X$};
    \node (W) at ({\df}, {-2*\dy}) {$\theta$};
    \node (Ys)at ({\df+\dx}, {-\dy}) {$\ss{Y_s}$};
    \node (Y) at ({\df+2*\dx}, {-1*\dy}) {$Y$};
    
    \draw[->] (X) -- (Ys);
    \draw[->] (W) -- (Ys);
    \draw[->] (Ys) -- (Y);
    \draw[->] (Ys) -- (Y);
  \end{tikzpicture}
  \caption{
    Graphical models under consideration.
    \textbf{A} The generative model for standard supervised learning with no data curation.
    \textbf{B} The generative model with data curation.
    \citep[Adapted with permission from][]{aitchison2020statistical}.
    \label{fig:graphical_model}
  }
\end{figure}
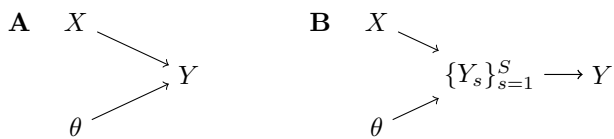

\cite{aitchison2020statistical} proposes a generative model of data curation that we will connect to the problem of OOD detection.
Given a random input, $X$, drawn from $P(X)$, a group of $S$ annotators (indexed by $s \in \{1, \dots, S \}$) are asked to assign labels $Y_s \in \mathcal{Y}$ to $X$, where $\mathcal{Y}=\{1,\dotsc,C\}$ represents the label set of $C$ classes.
If $X$ is OOD, annotators are instructed to label the image randomly. 
We assume that if the class-label is well-defined, sufficiently expert annotators will all agree on the label, so consensus is reached, $Y_1\ceq Y_2\ceq\dotsm\ceq Y_S$, and the image will be included in the dataset. 
Any disagreement is assumed to arise because the image is OOD, and such images are excluded from the dataset. 
In short, the final label $Y$ is chosen to be $Y_1$ if consensus was reached and $\texttt{Undef}$ otherwise (Fig.~\ref{fig:graphical_model}B).
\begin{align}
  Y | \ss{Y_s} &= \begin{cases}
    Y_1 & \text{if } Y_1\ceq Y_2\ceq \dotsm \ceq Y_S\\
    \texttt{Undef} & \text{otherwise}
  \end{cases}
\end{align}
From the equation above, we see that $Y \in \mathcal{Y} \cup \cb{\texttt{Undef}}$, that is, $Y$ could be any element from the label set $\mathcal{Y}$ if annotators come to agreement or $\texttt{Undef}$ if consensus is not reached. 
Suppose further that all annotators are IID (in the sense that their probability distribution over labels given an input image is the same).
Then, the probability of $Y\in \mathcal{Y}$ can be written as
\begin{align}
  \nonumber
  \P{Y\ceq y| X, \theta} &= \P{\ss{Y_s \ceq  y}| X, \theta} \\
  \nonumber
  &= {\textstyle\prod}_{s=1}^S \P{Y_s\ceq y| X, \theta}\\
  \label{eq:con_like}
  &= \P{Y_s\ceq y| X, \theta}^S = p_y^S(X)
\end{align}
where we have abbreviated the single-annotator probability as $p_y(X)=\P{Y_s\ceq y| X, \theta}$.
When consensus is not reached (noconsensus), we have:
\begin{align} 
  \nonumber
  \P{Y\ceq \texttt{Undef}| X, \theta} &=  1 - \sum_{y\in \mathcal{Y}} \P{Y\ceq y| X, \theta} \\
  \label{eq:noncon_like}
  &=  1-\sum_{y \in \mathcal{Y}} p_y^S(X).
\end{align}
Notice that the maximum of  Eq.~\eqref{eq:noncon_like} is achieved when the predictive distribution is uniform: $p_y = 1/C, \forall y \in \mathcal{Y}$, as can be shown using a Lagrange multiplier $\gamma$ to capture the normalization constraint,
\begin{align}
  L &= \b{1-\tsum_{y} p_{y}^S} + \gamma \b{1-\tsum_{y} p_{y}}\\
  0 &= \dd[L]{p_y} = - S p_y^{S-1} - \gamma
\end{align}
The value of $p_y$ with maximal $L$ is independent of $y$, so $p_y$ is the same for all $y\in\mathcal{Y}$, and we must therefore have $p_y=1/C$.
In addition, the minimum of zero is achieved when one of the $C$ classes has a probability $p_y(X) = 1$. 
Therefore, an input with high predictive (aleatoric) uncertainty is, \textit{by definition}, an input with a high probability of disagreement amongst multiple annotators, which corresponds to being OOD.

\section{Methods}
\begin{figure}[t]
  \centering
  \includegraphics[width=0.75\textwidth]{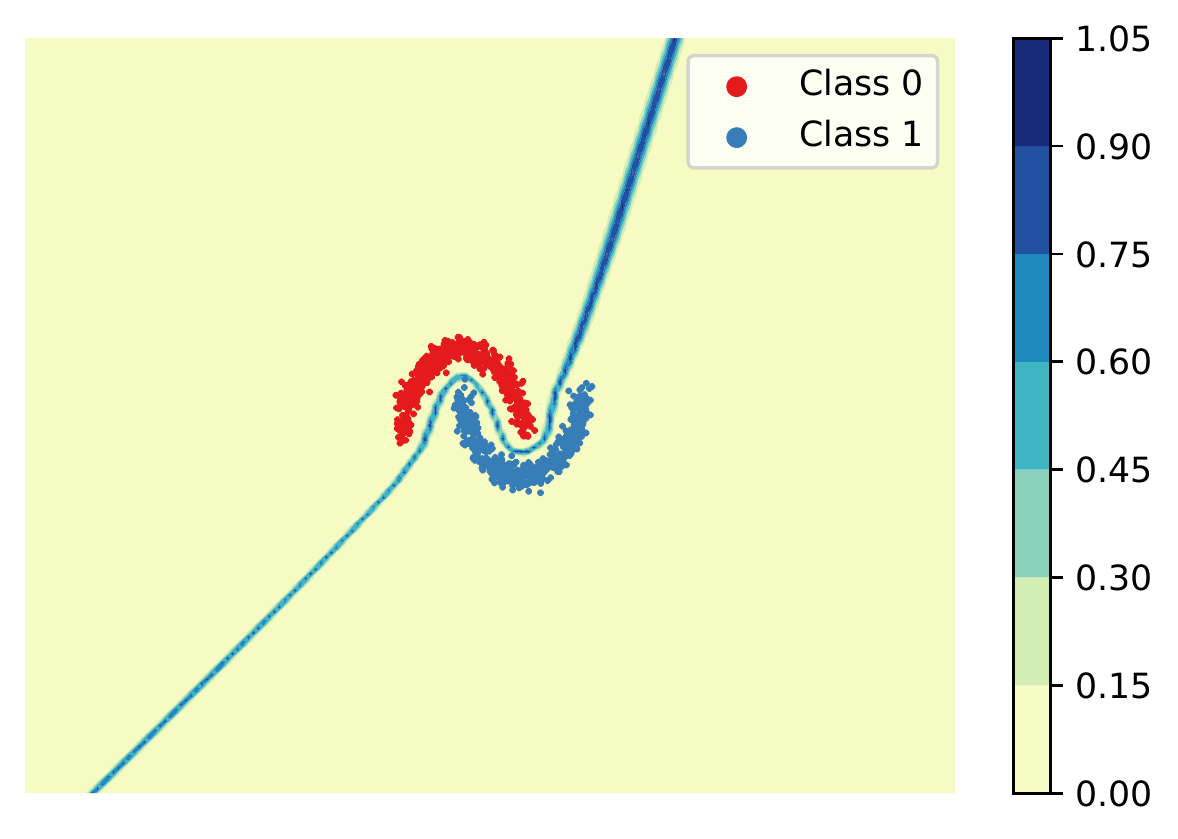}
  \caption{
  A visualisation of $\P{Y\ceq\texttt{Undef}| X, \theta}$ for toy data (red and blue points) for a network trained without inputs with undefined class-labels.}
    \label{fig:toy_data_baseline} 
\end{figure}

We are able to form a principled log-likelihood objective by combining Eq.~\eqref{eq:con_like} for inputs with a well-defined class-label (denoted by $\Din$) and Eq.~\eqref{eq:noncon_like} for OOD inputs without a well-defined class label (denoted by $\Dout$).
However, this model was initially developed for cold-posteriors \citep{aitchison2020statistical} and semi-supervised learning \citep{aitchison2021statistical} where the noconsensus inputs were not known and were omitted from the dataset.
In contrast, and following \citet{hendrycks2018deep}, we use proxy datasets for OOD inputs, and explicitly maximize the probability of inputs from those proxy datasets having being OOD (Eq.~\ref{eq:new:noncon_like}).
Importantly, now that we explicitly fit the probability of an undefined class-label, we need to introduce a little more flexibility into the model.
In particular, a key issue with the current model is that more annotators, $S$, implies a higher chance of disagreement hence implying more OOD images.
Thus, arbitrary choices about the relative amount of training data with well-defined and undefined class-labels might cause issues.
To avoid any such issues, we modify the undefined-class probability by including a base-rate or bias parameter, $c$, which modifies the log-odds for well-defined vs undefined class-labels.
In particular, we define the logits to be,
\begin{align}
  \ell_0 &= c + \log \b{1 - \tsum_{y\in\mathcal{Y}} p_y^S(X)}\\
  \ell_{y\in\mathcal{Y}} &= \log p_y^S(X)
\end{align}
where $p_y(X)$ is the single-annotator probability output by the neural network.
\begin{align}
  \label{eq:new:noncon_like}
  \P{Y\ceq\texttt{Undef}| X, \theta} &= \frac{e^{\ell_0}}{e^{\ell_0} + \sum_{y\in \mathcal{Y}} e^{\ell_y}} \\
  \label{eq:new:con_like}
  \P{Y\ceq y| X, \theta} &= \frac{e^{\ell_y}}{e^{\ell_0} + \sum_{y\in \mathcal{Y}} e^{\ell_y}}
\end{align}
with $c=0$, this reverts to Eq.~\eqref{eq:con_like} and \eqref{eq:noncon_like}, while non-zero $c$ allow us to modify the ratio of well-defined to undefined class-labels to match that in the training data.

Of course, we do not have the actual datapoints that were rejected during the data curation process, so instead $\Dout$ is a proxy dataset (e.g. taking CIFAR-10 as $\Din$, we might use downsampled ImageNet with 1000 classes as $\Dout$).
The objective is,
\begin{multline}
    \label{eq:obj}
    \mathcal{L} = \mathbb{E}_{\Din} \left[\log \P{Y\ceq y| X, \theta} \right] \\+
       \lambda \mathbb{E}_{\Dout} \left[\log  \P{Y\ceq\texttt{Undef}| X, \theta} \right]
\end{multline}
where $\lambda$ represents the relative quantity of inputs with undefined to well-defined class-labels.
We use $\lambda=1$ both for simplicity and because the inclusion of the bias parameter, $c$, should account for any mismatch between the ``true'' and proxy ratios of inputs with well-defined and undefined class-labels.
In addition, we use a fixed value of $S=10$ as is suggested by \cite{aitchison2020statistical} and we learn $c$ via backpropagation during training.

Lastly, since our objective is a well-defined likelihood function that jointly models $\Din$ and $\Dout$, we can easily turn our model into a fully Bayesian one by adding a prior distribution on the neural network parameters, $\theta$, and then perform approximate inference approaches (e.g. stochastic gradient Markov chain Monte Carlo) to estimate the posterior distribution over $\theta$.
The use of Bayesian inference in our approach allows us to incorporate both epistemic uncertainty and aleatoric uncertainty when detecting OOD samples and we will show in next section that combining two types of uncertainty together can lead to performance superior than using either of them alone.


\begin{figure*}[t]
  \centering
  \includegraphics[width=0.80 \textwidth]{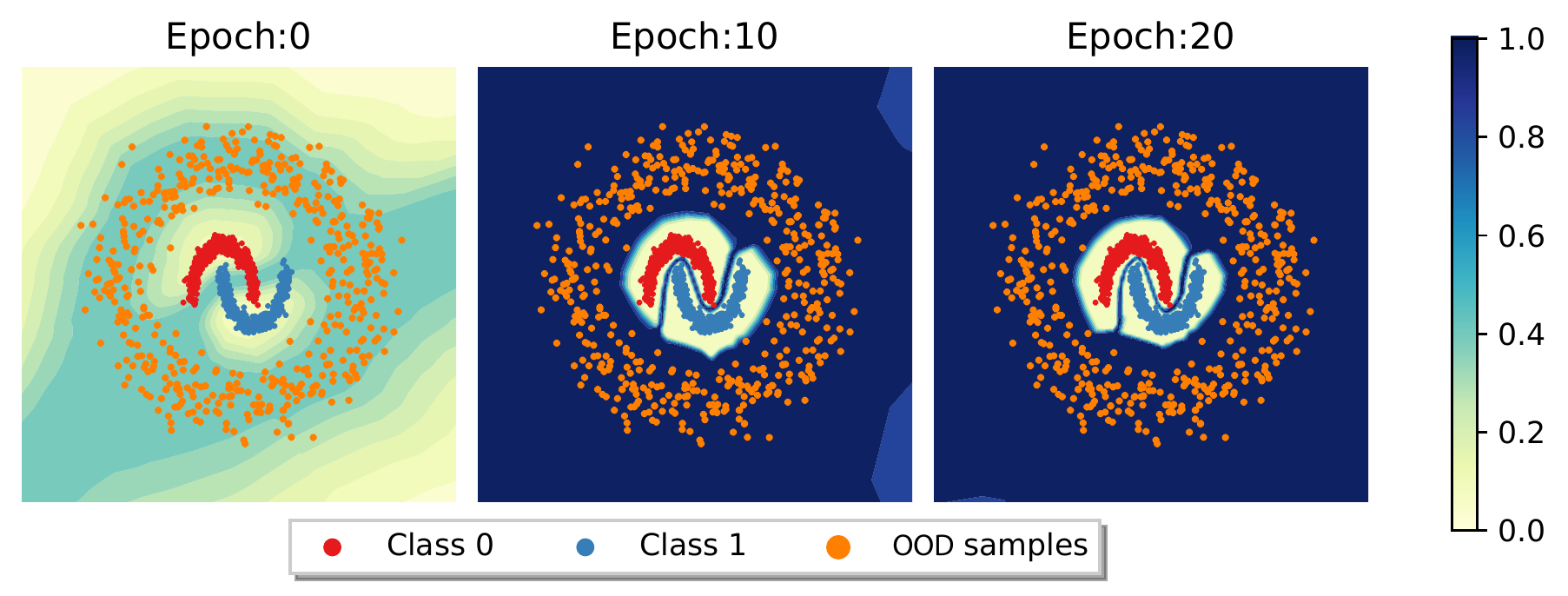}
  \caption{
  A visualisation of the dataset and the experiment results. Samples from the two classes are represented by red and blue points respectively. Orange points represent synthetic OOD inputs with undefined class-labels. The colormap shows $\P{Y\ceq\texttt{Undef}| X, \theta}$, with darker colors indicating more predictive uncertainty and higher undefined probability.
  }
    \label{fig:toy_data} 
\end{figure*}


\section{Results}

\subsection{Toy data}


First, we conducted an experiment on a toy dataset. 
We use the \texttt{make\_moons} generator from  Scikit-learn~\citep{scikit-learn} to generate labelled examples of two classes, with 2000 examples per class (Fig.~\ref{fig:toy_data_baseline}~and~\ref{fig:toy_data}).
For synthetic OOD points, we use 2000 samples generated from the \texttt{make\_circles} generator (Fig.~\ref{fig:toy_data}).

We trained a three-layer fully connected neural network with 32 hidden units in each hidden layer for 20 epochs using Adam \citep{kingma2015adam} with a constant learning rate of $5\times 10^{-3}$. 
Initially, we trained without data with undefined class-labels (Fig.~\ref{fig:toy_data_baseline}), and plotted the OOD probability, $\P{Y=\texttt{Undef}| X, \theta}$, with high undefined probabilities and uncertain classifications being denoted by darker colors.
This classifier is very certain for all inputs, and therefore judges there to be a well-defined class-label almost everywhere, except in a small region around the decision boundary.
In contrast, when training with OOD inputs (Fig.~\ref{fig:toy_data}), the model learns to take points around the training data as having a well-defined class-label (lighter colors), then classifies points further away as being OOD (darker colors).

\subsection{Image classification}

In this section, we demonstrate the effectiveness of our approach via large scale image classification experiments.
We considered two different baselines, in addition to our method.
First, in ``BNN'', we followed the usual OOD detection procedure for Bayesian neural networks in training on our in-distribution dataset, $\Din$, using SGLD, and ignored our OOD dataset, $\Dout$, as it is not clear how to incorporate these points into a classical Bayesian neural network.
Second, in ``OE'', we used the non-Bayesian method of \citet{hendrycks2018deep}, which trains on $\Din$, and incorporates an objective that encourages uncertainty on $\Dout$.
Our method uses a BNN, as in the BNN baseline, but additionally trains on $\Dout$ using the log-likelihood from Eq.~\eqref{eq:noncon_like} to increase uncertainty on those points.

\paragraph{Datasets}\label{sec:datasets}
We used CIFAR-10 and CIFAR-100 as $\Din$, and downsampled ImageNet as our training $\Dout$.
The Downsampled ImageNet dataset introduced by \cite{Oord2016PixelRN} is a downsampled version of ImageNet \citep{deng2009imagenet}\footnote{access governed agreement at \url{image-net.org/download.php}; underlying images are not owned by ImageNet and have a mixture of licenses} with the labels removed. The dataset is designed for unsupervised learning tasks such as density estimation and image generation.

At test time, we consider a number of OOD datasets that the model was not trained on, as proposed in \citet{hendrycks2016baseline} and as implemented by \citet{hendrycks2018deep}:

\begin{enumerate}[noitemsep,topsep=0pt, partopsep=0pt]
    \item Isotropic zero-mean Gaussian noise with $\sigma=0.5$
    \item Rademacher noise where each dimension is $-1$ or $1$ with equal probability.
    \item Blobs data of algorithmically generated amorphous shapes with definite edges
    \item Texture data~\citep{cimpoi14describing} of textural images in the wild.
    \item SVHN~\citep{Netzer2011ReadingDI} which contains 32x32 colour images of house numbers.
\end{enumerate}

\newcolumntype{B}{<{\hspace{-1.2ex}}c}
\begin{table*}[]
    \centering
    \begin{tabular}{llBBBBBB}
    
    \toprule
    \multicolumn{2}{c}{Dataset$\quad\quad\quad\quad$} &
      \multicolumn{3}{c}{AUROC ~$\uparrow$} & 
      \multicolumn{3}{c}{FPR95 ~$\downarrow$} \\
      \midrule
      {$\Din$} & {$\mathcal{D}_{\text{test}}$}& {BNN} & {OE}  & {Ours} &  {BNN} & {OE}  & {Ours} \\
      \midrule
      & Gaussian        &91.73$\pm$8.88     & 94.99$\pm$3.80 & \bf{100.00$\pm$0.0}
                        &13.91$\pm$11.05   & 9.03$\pm$5.83& \bf{0.00$\pm$0.0}\\
      & Rad.\           &93.78$\pm$3.90     & 95.96$\pm$1.70  & \bf{100.00$\pm$0.0} 
                        &11.01$\pm$7.05   & 7.62$\pm$2.49 & \bf{0.00$\pm$0.0}\\
      CIFAR-10 & Blob   &89.41$\pm$2.44     & 90.57$\pm$2.38 & \bf{100.00$\pm$0.0}
                        & 35.16$\pm$5.19   &33.25$\pm$10.17 & \bf{0.00$\pm$0.0}\\
      & Texture         & 89.39$\pm$0.49    & 88.21$\pm$1.16& \bf{95.25$\pm$0.63}
                        & 37.61$\pm$1.14   &52.17$\pm$6.15  & \bf{25.19$\pm$3.93}\\
      & SVHN            & 90.42$\pm$1.83    & 93.33$\pm$0.89 & \bf{97.27$\pm$0.52}
                        & 28.55$\pm$4.36   & 19.43$\pm$2.46 & \bf{12.28$\pm$2.28}\\
      \midrule
      & Gaussian & 93.16$\pm$3.38 & 80.10$\pm$8.97&  \bf{100.00$\pm$0.00 }
      & 14.47$\pm$5.45 & 32.92$\pm$12.43&  \bf{0.00$\pm$0.00 } \\
& Rad.\ & 85.39$\pm$6.51 & 95.34$\pm$5.69&  \bf{100.00$\pm$0.00 }
& 26.60$\pm$11.01 & 10.95$\pm$8.35&  \bf{0.00$\pm$0.00 } \\
  CIFAR-100 & Blob & 89.15$\pm$1.79 & 87.93$\pm$7.28&  \bf{100.00$\pm$0.00 }
  & 29.95$\pm$3.59 & 36.62$\pm$16.30&  \bf{0.01$\pm$0.01 } \\
& Texture & 75.97$\pm$0.52 & 75.15$\pm$1.05&  \bf{85.07$\pm$0.47 }
& 69.74$\pm$2.27 & 76.72$\pm$2.74&  \bf{58.43$\pm$2.09 } \\
& SVHN & 81.04$\pm$0.73 & 77.85$\pm$2.20&  \bf{89.66$\pm$1.77 }
& 52.18$\pm$2.90 & 63.67$\pm$6.24&  \bf{42.70$\pm$5.57 } \\
      \bottomrule
    \end{tabular}
    \caption{Experimental results on a range of different datasets for AUROC and FPR95.  Note the arrows indicate the ``better'' direction (e.g.\ so higher AUROC is better). $\Din$ represents the in-distribution dataset. $\mathcal{D}_{\text{test}}$ is the testing out-of-distribution dataset. (The results reported are mean and standard error computed over 6 runs of different random seeds.) }
    \label{table:img_result}
\end{table*}

\paragraph{OOD score}
At test time, to distinguish between OOD and in-distribution examples, we need a score that measures the model's uncertainty.
There are several model-specific choices.
In our model one can choose to use the OOD probability (Eq.~\ref{eq:noncon_like}) as the score. 
\cite{hendrycks2016baseline, hendrycks2018deep} use the negative maximum softmax probability~\citep{hendrycks2016baseline}. 
However, to ensure a fair comparison, we used one metric that makes sense for all methods considered, the total uncertainty~\citep{depeweg2018decomposition}. 
The total uncertainty is the entropy of the predictive distribution, marginalising over uncertainty in the neural network parameters, $\mathbb{H}\left[p(y \mid x^*)\right]$, which equals the sum of aleatoric uncertainty and epistemic uncertainty. 
Note that the total uncertainty from OE only contains aleatoric uncertainty since the model is fully deterministic.
In contrast, the standard BNN and our BNN approach both have aleatoric and epistemic uncertainty.
The key differences is that in our model, the aleatoric uncertainty is shaped by outlier exposure, whereas in a standard BNN, the aleatoric uncertainty is determined solely by the in-distribution data.

\paragraph{Evaluation metrics} \label{sec:metric}
%
OOD detection is in essence a binary classification problem. 

It is therefore sensible to use metrics for binary classification to evaluate a model's ability to detect OOD inputs.
%

As such, to allow for comparison to past work we use two metrics.
%
%
%
First, the area under the receiver operating characteristic curve (AUROC) indicates how well the model can discriminate between positive and negative classes, for example, a model with AUROC of $0.9$ would assign a higher probability to an input with undefined class-label than one with a well-defined class-label $90\%$ of the time.
A perfect model would have an AUROC of $1.0$ while an uninformative model with random outputs would have an AUROC of $0.5$.

Secondly, the false positive rate at $N\%$ true positive rate (FPR$N$) computes the probability of an input being misclassified as having an undefined class-label (false positive) when at least $N\%$ of the true inputs with undefined class-labels are correctly detected (true positive). 
In practice, we would like to have a model with low FPR$N\%$ since an ideal model should detect nearly all inputs with undefined class-labels while raising as few false alarms as possible. In our experiments, we let $N=95$.

%
%

\paragraph{Networks and training protocols}
As is mentioned above, our experiments involve comparison among outlier exposure (OE), Bayesian neural network (BNN) and our proposed approach (Ours).

The network architecture is chosen to be a 40-2 Wide Residual Network ~\citep{Zagoruyko2016WRN} for all experiments.
OE was trained directly using the code from \citet{hendrycks2018deep}.

For BNN and our approach, we used Cyclical Stochastic Gradient MCMC~\citep{Zhang2020Cyclical} to perform approximate inference over the network parameters. The implementation and learning rate scheduling largely followed \cite{Zhang2020Cyclical}. In particular, the cycle length was chosen to be 50 epochs and we ran 4 cycles in total. At each cycle, we stored weights at the last 3 epochs as samples from the approximate posterior. In addition, we used a temperature of 0.1 on the likelihood for the baseline BNN, so as to match $S=10$ in the labelled likelihood for our model \citep[Eq.~\ref{eq:con_like}][]{aitchison2021statistical}.
For all experiments, at each iteration, we feed the network with a mini-batch made up of 128 samples from $\Din$ and 256 samples from $\Dout$ such that $\Dout$ will not usually be entirely traversed during a training epoch.
%

\paragraph{Results}
Broadly, we found that our approach gave superior performance to the OE and BNN using AUROC and FPR95 (Table~\ref{table:img_result}).

\section{Related work}
\label{sec:related_work}

The most closely related prior work is that on OOD detection using epistemic uncertainty.
The history of this idea is a bit difficult to trace, as the idea evolved rapidly into benchmark for Bayesian neural networks \citep{lakshminarayanan2017simple,louizos2017multiplicative,pawlowski2017implicit,krueger2017bayesian,henning2018approximating,maddox2019simple,ovadia2019can,fortuin2021bayesian}, with very few papers written directly about OOD detection using BNNs \citep[though see][]{malinin2018predictive}.

At the same time, a variety of approaches based on aleatoric uncertainty were developed.
This line of work started with \citet{hendrycks2016baseline}, who used the maximum softmax probability for a pre-trained network to identify OOD points.
Better separation between in and out of distribution points was obtained by \citet{liang2017principled} by temperature scaling and perturbing the inputs, and Probably Approximately Correct (PAC) guarantees were obtained by \citet{liu2018open}.
Later work trained on OOD data using an objective such as the cross entropy to the uniform distribution \citep{lee2017training,hendrycks2018deep,dhamija2018reducing}.
The requisite OOD data was either explicitly provided \citep{hendrycks2018deep} or obtained using a generative model \citep{lee2017training}.
In contrast, our method provides a formal account of aleatoric uncertainty based OOD detection as maximizing a likelihood, and integrates it with Bayesian epistemic uncertainty.

Compared with utilising a model's predictive uncertainty, a perhaps more straightforward approach is to train a binary classifier to perform OOD detection.
The required OOD inputs can be obtained from a proxy OOD dataset \citep{devries2018learning,hendrycks2018deep}, by taking misclassified points to be OOD \citep{devries2018learning}, or can be generated from a generative model \citep{vernekar2019out,guenais2020bacoun}.
However, approaches based on predictive uncertainty are generally found to perform better than those based on explicitly classifying OOD points \citep{hendrycks2018deep}. 
This might be expected as classification based approaches rely on the presence of an OOD dataset, whereas approaches based on predictive uncertainty \citep{hendrycks2016baseline} are able to extract useful information even in the absence of an outlier dataset.

There is also work on ``classification with rejection'', in high-risk settings such as medical decision making, where there is a very high cost for misclassifying an image (e.g.\ failing to diagnose cancer could lead to death). 
As such, the machine learning system should defer to a human expert whenever it is uncertain (perhaps because the model capacity is limited, it has not seen enough data, or the medical expert has side-information) \citep[e.g.][]{herbei2006classification,bartlett2008classification,cortes2016learning, mozannar2020consistent}.
Importantly, in classification with rejection there is a well-defined right answer for the rejected inputs, and we defer to the human expert because it is critically important to get that right answer. 
In contrast, in our setting, there is no meaningful right answer for the rejected inputs: (e.g.\ classify an image of a radio as cat vs dog).
Therefore, an expert would not be able to classify our rejected points because the corresponding ground-truth class-label is fundamentally ill-defined, and if forced to make a classification, all they could do is to choose a random label.

Finally, it is possible to solve the OOD detection task by fitting a generative model of inputs, and declaring inputs with low probability under that model as OOD \citep{wang2017safer,pidhorskyi2018generative}.
However, there are considerable subtleties in getting this approach to work well \citep{nalisnick2018deep,choi2018waic,shafaei2018does,hendrycks2018deep,ren2019likelihood,morningstar2020density}, as having a high probability under the probabilitistic generative model does not necessarily mean than an input point is typical of the dataset.
The generative approach is often much more difficult to apply in practice, simply due to the much greater difficulty in training good generative models, as compared with training classifiers.

\section{Conclusion}
We developed a likelihood for UCL points (a subset of OOD points), and used it to integrate OE methods within principled Bayesian inference.
The resulting Bayesian OE method gave superior performance to other methods, including pure aleatoric uncertainty and Bayesian methods without OE.


\newpage

\bibliographystyle{icml2020}
\bibliography{main}

\begin{thebibliography}{49}
\providecommand{\natexlab}[1]{#1}
\providecommand{\url}[1]{\texttt{#1}}
\expandafter\ifx\csname urlstyle\endcsname\relax
  \providecommand{\doi}[1]{doi: #1}\else
  \providecommand{\doi}{doi: \begingroup \urlstyle{rm}\Url}\fi

\bibitem[Aitchison(2020)]{aitchison2021statistical}
Aitchison, L.
\newblock A statistical theory of semi-supervised learning.
\newblock \emph{arXiv preprint arXiv:2008.05913}, 2020.

\bibitem[Aitchison(2021)]{aitchison2020statistical}
Aitchison, L.
\newblock A statistical theory of cold posteriors in deep neural networks.
\newblock In \emph{International Conference on Learning Representations}, 2021.
\newblock URL \url{https://openreview.net/forum?id=Rd138pWXMvG}.

\bibitem[Bartlett \& Wegkamp(2008)Bartlett and
  Wegkamp]{bartlett2008classification}
Bartlett, P.~L. and Wegkamp, M.~H.
\newblock Classification with a reject option using a hinge loss.
\newblock \emph{Journal of Machine Learning Research}, 9\penalty0 (8), 2008.

\bibitem[Choi et~al.(2018)Choi, Jang, and Alemi]{choi2018waic}
Choi, H., Jang, E., and Alemi, A.~A.
\newblock Waic, but why? generative ensembles for robust anomaly detection.
\newblock \emph{arXiv preprint arXiv:1810.01392}, 2018.

\bibitem[Cimpoi et~al.(2014)Cimpoi, Maji, Kokkinos, Mohamed, , and
  Vedaldi]{cimpoi14describing}
Cimpoi, M., Maji, S., Kokkinos, I., Mohamed, S., , and Vedaldi, A.
\newblock Describing textures in the wild.
\newblock In \emph{Proceedings of the {IEEE} Conf. on Computer Vision and
  Pattern Recognition ({CVPR})}, 2014.

\bibitem[Cortes et~al.(2016)Cortes, DeSalvo, and Mohri]{cortes2016learning}
Cortes, C., DeSalvo, G., and Mohri, M.
\newblock Learning with rejection.
\newblock In \emph{International Conference on Algorithmic Learning Theory},
  pp.\  67--82. Springer, 2016.

\bibitem[Deng et~al.(2009{\natexlab{a}})Deng, Dong, Socher, Li, Li, and
  Fei-Fei]{deng2009imagenet}
Deng, J., Dong, W., Socher, R., Li, L.-J., Li, K., and Fei-Fei, L.
\newblock Imagenet: A large-scale hierarchical image database.
\newblock In \emph{2009 IEEE conference on computer vision and pattern
  recognition}, pp.\  248--255. Ieee, 2009{\natexlab{a}}.

\bibitem[Deng et~al.(2009{\natexlab{b}})Deng, Dong, Socher, Li, Li, and
  Fei-Fei]{imagenet_cvpr09}
Deng, J., Dong, W., Socher, R., Li, L.-J., Li, K., and Fei-Fei, L.
\newblock {ImageNet: A Large-Scale Hierarchical Image Database}.
\newblock In \emph{CVPR09}, 2009{\natexlab{b}}.

\bibitem[Depeweg et~al.(2018)Depeweg, Hernandez-Lobato, Doshi-Velez, and
  Udluft]{depeweg2018decomposition}
Depeweg, S., Hernandez-Lobato, J.-M., Doshi-Velez, F., and Udluft, S.
\newblock Decomposition of uncertainty in bayesian deep learning for efficient
  and risk-sensitive learning.
\newblock In \emph{International Conference on Machine Learning}, pp.\
  1184--1193. PMLR, 2018.

\bibitem[Der~Kiureghian \& Ditlevsen(2009)Der~Kiureghian and
  Ditlevsen]{der2009aleatory}
Der~Kiureghian, A. and Ditlevsen, O.
\newblock Aleatory or epistemic? does it matter?
\newblock \emph{Structural safety}, 31\penalty0 (2):\penalty0 105--112, 2009.

\bibitem[DeVries \& Taylor(2018)DeVries and Taylor]{devries2018learning}
DeVries, T. and Taylor, G.~W.
\newblock Learning confidence for out-of-distribution detection in neural
  networks.
\newblock \emph{arXiv preprint arXiv:1802.04865}, 2018.

\bibitem[Dhamija et~al.(2018)Dhamija, G{\"u}nther, and
  Boult]{dhamija2018reducing}
Dhamija, A.~R., G{\"u}nther, M., and Boult, T.~E.
\newblock Reducing network agnostophobia.
\newblock \emph{arXiv preprint arXiv:1811.04110}, 2018.

\bibitem[Fortuin et~al.(2021)Fortuin, Garriga-Alonso, Wenzel, R{\"a}tsch,
  Turner, van~der Wilk, and Aitchison]{fortuin2021bayesian}
Fortuin, V., Garriga-Alonso, A., Wenzel, F., R{\"a}tsch, G., Turner, R.,
  van~der Wilk, M., and Aitchison, L.
\newblock Bayesian neural network priors revisited.
\newblock \emph{arXiv preprint arXiv:2102.06571}, 2021.

\bibitem[Fox \& {\"U}lk{\"u}men(2011)Fox and
  {\"U}lk{\"u}men]{fox2011distinguishing}
Fox, C.~R. and {\"U}lk{\"u}men, G.
\newblock Distinguishing two dimensions of uncertainty.
\newblock \emph{Essays in Judgment and Decision Making}, 2011.

\bibitem[Gu{\'e}nais et~al.(2020)Gu{\'e}nais, Vamvourellis, Yacoby,
  Doshi-Velez, and Pan]{guenais2020bacoun}
Gu{\'e}nais, T., Vamvourellis, D., Yacoby, Y., Doshi-Velez, F., and Pan, W.
\newblock Bacoun: Bayesian classifers with out-of-distribution uncertainty.
\newblock \emph{arXiv preprint arXiv:2007.06096}, 2020.

\bibitem[Hendrycks \& Gimpel(2016)Hendrycks and Gimpel]{hendrycks2016baseline}
Hendrycks, D. and Gimpel, K.
\newblock A baseline for detecting misclassified and out-of-distribution
  examples in neural networks.
\newblock \emph{arXiv preprint arXiv:1610.02136}, 2016.

\bibitem[Hendrycks et~al.(2018)Hendrycks, Mazeika, and
  Dietterich]{hendrycks2018deep}
Hendrycks, D., Mazeika, M., and Dietterich, T.
\newblock Deep anomaly detection with outlier exposure.
\newblock In \emph{International Conference on Learning Representations}, 2018.

\bibitem[Henning et~al.(2018)Henning, von Oswald, Sacramento, Surace, Pfister,
  and Grewe]{henning2018approximating}
Henning, C., von Oswald, J., Sacramento, J., Surace, S.~C., Pfister, J.-P., and
  Grewe, B.~F.
\newblock Approximating the predictive distribution via adversarially-trained
  hypernetworks.
\newblock \emph{Bayesian deep learning workshop at NeurIPS}, 2018.

\bibitem[Herbei \& Wegkamp(2006)Herbei and Wegkamp]{herbei2006classification}
Herbei, R. and Wegkamp, M.~H.
\newblock Classification with reject option.
\newblock \emph{The Canadian Journal of Statistics/La Revue Canadienne de
  Statistique}, pp.\  709--721, 2006.

\bibitem[Izmailov et~al.(2021)Izmailov, Vikram, Hoffman, and
  Wilson]{izmailov2021bayesian}
Izmailov, P., Vikram, S., Hoffman, M.~D., and Wilson, A.~G.
\newblock What are bayesian neural network posteriors really like?
\newblock In \emph{{ICML}}, volume 139 of \emph{Proceedings of Machine Learning
  Research}, pp.\  4629--4640. {PMLR}, 2021.

\bibitem[Kendall \& Gal(2017)Kendall and Gal]{kendall2017what}
Kendall, A. and Gal, Y.
\newblock What uncertainties do we need in bayesian deep learning for computer
  vision?
\newblock In \emph{{NIPS}}, pp.\  5574--5584, 2017.

\bibitem[Kingma \& Ba(2015)Kingma and Ba]{kingma2015adam}
Kingma, D.~P. and Ba, J.
\newblock Adam: A method for stochastic optimization.
\newblock In \emph{International Conference on Learning Representations}, 2015.

\bibitem[Krizhevsky et~al.(2009)Krizhevsky, Hinton,
  et~al.]{krizhevsky2009learning}
Krizhevsky, A., Hinton, G., et~al.
\newblock Learning multiple layers of features from tiny images.
\newblock \emph{Tech. report}, 2009.

\bibitem[Krueger et~al.(2017)Krueger, Huang, Islam, Turner, Lacoste, and
  Courville]{krueger2017bayesian}
Krueger, D., Huang, C., Islam, R., Turner, R., Lacoste, A., and Courville,
  A.~C.
\newblock Bayesian hypernetworks.
\newblock \emph{CoRR}, abs/1710.04759, 2017.

\bibitem[Lakshminarayanan et~al.(2017)Lakshminarayanan, Pritzel, and
  Blundell]{lakshminarayanan2017simple}
Lakshminarayanan, B., Pritzel, A., and Blundell, C.
\newblock Simple and scalable predictive uncertainty estimation using deep
  ensembles.
\newblock In \emph{{NIPS}}, pp.\  6402--6413, 2017.

\bibitem[Lee et~al.(2017)Lee, Lee, Lee, and Shin]{lee2017training}
Lee, K., Lee, H., Lee, K., and Shin, J.
\newblock Training confidence-calibrated classifiers for detecting
  out-of-distribution samples.
\newblock \emph{arXiv preprint arXiv:1711.09325}, 2017.

\bibitem[Liang et~al.(2017)Liang, Li, and Srikant]{liang2017principled}
Liang, S., Li, Y., and Srikant, R.
\newblock Principled detection of out-of-distribution examples in neural
  networks.
\newblock \emph{arXiv preprint arXiv:1706.02690}, pp.\  655--662, 2017.

\bibitem[Liu et~al.(2018)Liu, Garrepalli, Dietterich, Fern, and
  Hendrycks]{liu2018open}
Liu, S., Garrepalli, R., Dietterich, T., Fern, A., and Hendrycks, D.
\newblock Open category detection with pac guarantees.
\newblock In \emph{International Conference on Machine Learning}, pp.\
  3169--3178. PMLR, 2018.

\bibitem[Louizos \& Welling(2017)Louizos and
  Welling]{louizos2017multiplicative}
Louizos, C. and Welling, M.
\newblock Multiplicative normalizing flows for variational bayesian neural
  networks.
\newblock In \emph{{ICML}}, volume~70 of \emph{Proceedings of Machine Learning
  Research}, pp.\  2218--2227. {PMLR}, 2017.

\bibitem[Maddox et~al.(2019)Maddox, Izmailov, Garipov, Vetrov, and
  Wilson]{maddox2019simple}
Maddox, W.~J., Izmailov, P., Garipov, T., Vetrov, D.~P., and Wilson, A.~G.
\newblock A simple baseline for bayesian uncertainty in deep learning.
\newblock \emph{NeurIPS}, pp.\  13132--13143, 2019.

\bibitem[Malinin \& Gales(2018)Malinin and Gales]{malinin2018predictive}
Malinin, A. and Gales, M.
\newblock Predictive uncertainty estimation via prior networks.
\newblock \emph{arXiv preprint arXiv:1802.10501}, 2018.

\bibitem[Malinin et~al.(2020)Malinin, Mlodozeniec, and
  Gales]{malinin2020ensemble}
Malinin, A., Mlodozeniec, B., and Gales, M.
\newblock Ensemble distribution distillation.
\newblock In \emph{International Conference on Learning Representations}, 2020.
\newblock URL \url{https://openreview.net/forum?id=BygSP6Vtvr}.

\bibitem[Morningstar et~al.(2020)Morningstar, Ham, Gallagher, Lakshminarayanan,
  Alemi, and Dillon]{morningstar2020density}
Morningstar, W.~R., Ham, C., Gallagher, A.~G., Lakshminarayanan, B., Alemi,
  A.~A., and Dillon, J.~V.
\newblock Density of states estimation for out-of-distribution detection.
\newblock \emph{arXiv preprint arXiv:2006.09273}, 2020.

\bibitem[Mozannar \& Sontag(2020)Mozannar and Sontag]{mozannar2020consistent}
Mozannar, H. and Sontag, D.
\newblock Consistent estimators for learning to defer to an expert.
\newblock In \emph{International Conference on Machine Learning}, pp.\
  7076--7087. PMLR, 2020.

\bibitem[Nalisnick et~al.(2018)Nalisnick, Matsukawa, Teh, Gorur, and
  Lakshminarayanan]{nalisnick2018deep}
Nalisnick, E., Matsukawa, A., Teh, Y.~W., Gorur, D., and Lakshminarayanan, B.
\newblock Do deep generative models know what they don't know?
\newblock \emph{arXiv preprint arXiv:1810.09136}, 2018.

\bibitem[Netzer et~al.(2011)Netzer, Wang, Coates, Bissacco, Wu, and
  Ng]{Netzer2011ReadingDI}
Netzer, Y., Wang, T., Coates, A., Bissacco, A., Wu, B., and Ng, A.
\newblock Reading digits in natural images with unsupervised feature learning.
\newblock \emph{Feature Learning NIPS Workshop on Deep Learning and
  Unsupervised Feature Learning}, 2011.

\bibitem[Oord et~al.(2016)Oord, Kalchbrenner, and Kavukcuoglu]{Oord2016PixelRN}
Oord, A., Kalchbrenner, N., and Kavukcuoglu, K.
\newblock Pixel recurrent neural networks.
\newblock \emph{ArXiv}, abs/1601.06759, 2016.

\bibitem[Ovadia et~al.(2019)Ovadia, Fertig, Ren, Nado, Sculley, Nowozin,
  Dillon, Lakshminarayanan, and Snoek]{ovadia2019can}
Ovadia, Y., Fertig, E., Ren, J., Nado, Z., Sculley, D., Nowozin, S., Dillon,
  J., Lakshminarayanan, B., and Snoek, J.
\newblock Can you trust your model's uncertainty? evaluating predictive
  uncertainty under dataset shift.
\newblock \emph{Advances in Neural Information Processing Systems},
  32:\penalty0 13991--14002, 2019.

\bibitem[Pawlowski et~al.(2017)Pawlowski, Rajchl, and
  Glocker]{pawlowski2017implicit}
Pawlowski, N., Rajchl, M., and Glocker, B.
\newblock Implicit weight uncertainty in neural networks.
\newblock \emph{CoRR}, abs/1711.01297, 2017.

\bibitem[Pedregosa et~al.(2011)Pedregosa, Varoquaux, Gramfort, Michel, Thirion,
  Grisel, Blondel, Prettenhofer, Weiss, Dubourg, Vanderplas, Passos,
  Cournapeau, Brucher, Perrot, and Duchesnay]{scikit-learn}
Pedregosa, F., Varoquaux, G., Gramfort, A., Michel, V., Thirion, B., Grisel,
  O., Blondel, M., Prettenhofer, P., Weiss, R., Dubourg, V., Vanderplas, J.,
  Passos, A., Cournapeau, D., Brucher, M., Perrot, M., and Duchesnay, E.
\newblock Scikit-learn: Machine learning in {P}ython.
\newblock \emph{Journal of Machine Learning Research}, 12:\penalty0 2825--2830,
  2011.

\bibitem[Pidhorskyi et~al.(2018)Pidhorskyi, Almohsen, Adjeroh, and
  Doretto]{pidhorskyi2018generative}
Pidhorskyi, S., Almohsen, R., Adjeroh, D.~A., and Doretto, G.
\newblock Generative probabilistic novelty detection with adversarial
  autoencoders.
\newblock \emph{arXiv preprint arXiv:1807.02588}, 2018.

\bibitem[Postels et~al.(2020)Postels, Blum, Str{\"u}mpler, Cadena, Siegwart,
  Van~Gool, and Tombari]{postels2020hidden}
Postels, J., Blum, H., Str{\"u}mpler, Y., Cadena, C., Siegwart, R., Van~Gool,
  L., and Tombari, F.
\newblock The hidden uncertainty in a neural networks activations.
\newblock \emph{arXiv preprint arXiv:2012.03082}, 2020.

\bibitem[Ren et~al.(2019)Ren, Liu, Fertig, Snoek, Poplin, DePristo, Dillon, and
  Lakshminarayanan]{ren2019likelihood}
Ren, J., Liu, P.~J., Fertig, E., Snoek, J., Poplin, R., DePristo, M.~A.,
  Dillon, J.~V., and Lakshminarayanan, B.
\newblock Likelihood ratios for out-of-distribution detection.
\newblock \emph{arXiv preprint arXiv:1906.02845}, 2019.

\bibitem[Shafaei et~al.(2018)Shafaei, Schmidt, and Little]{shafaei2018does}
Shafaei, A., Schmidt, M., and Little, J.~J.
\newblock Does your model know the digit 6 is not a cat? a less biased
  evaluation of" outlier" detectors.
\newblock \emph{CoRR}, 2018.

\bibitem[Vernekar et~al.(2019)Vernekar, Gaurav, Abdelzad, Denouden, Salay, and
  Czarnecki]{vernekar2019out}
Vernekar, S., Gaurav, A., Abdelzad, V., Denouden, T., Salay, R., and Czarnecki,
  K.
\newblock Out-of-distribution detection in classifiers via generation.
\newblock \emph{arXiv preprint arXiv:1910.04241}, 2019.

\bibitem[Wang et~al.(2017)Wang, Wang, Tamar, Chen, and Abbeel]{wang2017safer}
Wang, W., Wang, A., Tamar, A., Chen, X., and Abbeel, P.
\newblock Safer classification by synthesis.
\newblock \emph{arXiv preprint arXiv:1711.08534}, 2017.

\bibitem[Wen et~al.(2019)Wen, Tran, and Ba]{wen2019batchensemble}
Wen, Y., Tran, D., and Ba, J.
\newblock Batchensemble: an alternative approach to efficient ensemble and
  lifelong learning.
\newblock In \emph{International Conference on Learning Representations}, 2019.

\bibitem[Zagoruyko \& Komodakis(2016)Zagoruyko and Komodakis]{Zagoruyko2016WRN}
Zagoruyko, S. and Komodakis, N.
\newblock Wide residual networks.
\newblock In \emph{BMVC}, 2016.

\bibitem[Zhang et~al.(2020)Zhang, Li, Zhang, Chen, and
  Wilson]{Zhang2020Cyclical}
Zhang, R., Li, C., Zhang, J., Chen, C., and Wilson, A.~G.
\newblock Cyclical stochastic gradient mcmc for bayesian deep learning.
\newblock In \emph{International Conference on Learning Representations}, 2020.
\newblock URL \url{https://openreview.net/forum?id=rkeS1RVtPS}.

\end{thebibliography}

\end{document}